\pdfoutput=1

\documentclass[11pt, dvipsnames, table]{article}

\usepackage[]{acl}

\usepackage{times}
\usepackage{latexsym}
\usepackage[T1]{fontenc}
\usepackage[utf8]{inputenc}
\usepackage{microtype}
\usepackage{inconsolata}
\usepackage{amsmath}
\usepackage{amssymb}
\usepackage{amsthm}
\usepackage{graphicx}
\usepackage{mathtools}
\usepackage{listings}
\usepackage{tikz}
\usepackage{siunitx}
\usepackage{tipa}
\usepackage{float}
\usepackage{bbm}
\usepackage{breqn}
\usepackage{utils}
\usepackage{tikz-dependency}

\usepackage{caption}
\usepackage{subcaption}
\usepackage{linguex}
\usepackage{xcolor}
\usepackage{booktabs}

\usepackage[textsize=scriptsize]{todonotes}

\makeatletter
\newcommand*\iftodonotes{\if@todonotes@disabled\expandafter\@secondoftwo\else\expandafter\@firstoftwo\fi}  
\makeatother

\DeclareMathOperator*{\argmax}{argmax}

\usepackage{xspace}
\newcommand{\just}{\textit{just}\xspace}
\newcommand{\justs}{\textit{just}s\xspace}

\definecolor{blu}{HTML}{3468C0}
\definecolor{yello}{HTML}{e6ab02}
\definecolor{piink}{HTML}{e7298a}
\definecolor{greeen}{HTML}{66a61e}
\definecolor{reed}{HTML}{b2182b}
\definecolor{firebrick}{HTML}{B22222}
\definecolor{steelblue}{HTML}{4682B4}
\definecolor{puuurpl}{HTML}{B5739D}
\definecolor{gren}{HTML}{1b9e77}
\definecolor{prpl}{HTML}{7570b3}
\definecolor{orang}{HTML}{d95f02}
\definecolor{vlightgreen}{HTML}{82B366}
\definecolor{vviolet}{HTML}{56517E}

\definecolor{regexv1}{HTML}{B46504}
\definecolor{regexv2}{HTML}{56517E}
\definecolor{regexv3}{HTML}{0E8088}

\definecolor{naanpollution}{HTML}{488795}
\definecolor{ananpollution}{HTML}{ae4d9f}

\usepackage{booktabs, multirow}

\Crefname{figure}{{Fig.}}{{Figs.}}
\crefname{section}{§}{§§}
\Crefname{section}{§}{§§}
\Crefname{appendix}{{App.}}{{Apps.}}

\title{Is It \emph{JUST} Semantics? \\ A Case Study of Discourse Particle Understanding in LLMs}

\author{William Sheffield$^{1}$\ \ \
Kanishka Misra$^{2,\bigstar}$\ \ \
Valentina Pyatkin$^{3,4}$\\
\bf
Ashwini Deo$^{1}$\ \ \
Kyle Mahowald$^{1}$\ \ \
Junyi Jessy Li$^{1}$
\\
$^1$Linguistics, The University of Texas at Austin\ \ \
$^2$Toyota Technological Institute at Chicago\\
$^3$Allen Institute for AI\ \ \
$^4$University of Washington\\
{\tt \{sheffieldw, mahowald, jessy\}@utexas.edu}\ \ \
{\tt kanishka@ttic.edu} \ \ \
\\
}

\begin{document}

\maketitle

\begin{abstract}
Discourse particles are crucial elements that subtly shape the meaning of text.  These words, often polyfunctional, 
give rise to nuanced and often quite disparate semantic/discourse effects,
as exemplified by the diverse uses of the particle \just (e.g., exclusive, temporal, emphatic).  
This work investigates the capacity of LLMs to distinguish the fine-grained senses of English \just, a well-studied example in formal semantics, using data meticulously created and labeled by expert linguists.
Our findings reveal that while LLMs exhibit some ability to differentiate between broader categories, they struggle to fully capture more subtle nuances, highlighting a gap in their understanding of discourse particles.
\end{abstract}

{\let\thefootnote\relax\footnotetext{\hspace{-0.2cm}$^{\bigstar}$Work partly done at UT-Austin before joining TTIC.}}

\section{Introduction}
\label{sec:intro}
Discourse particles are words that comment on aspects of the discourse context or interlocutor attitudes, giving rise to  discourse effects that are often difficult to pin down. 
In some of their uses, their contribution is straightforward.
For example, the \emph{just} in ``\emph{Betsy just eats chicken nuggets}'' tells us that chicken nuggets are the only thing Betsy eats. 
Without the \just, we learn nothing about the other things Betsy will (not) eat.
But not all uses of a polyfunctional discourse particle are easily unifiable: consider the occurrences of \just in ``\emph{My brother \textbf{just} flew in to town}'' (\just $\approx$ recently) and ``\emph{I \textbf{just} won't stand for this injustice}'' (\just $\approx$ simply), or the latter two in ``\textit{A just judge \textbf{just} wouldn't stand for the laws \textbf{just} passed}''.

From the view of formal semantics, these particles are difficult to analyze, partly because of their rich diversity of senses bundled into one word and partly because of the difficulty of characterizing each individual use \cite{lee1987semantics, bonomi1993only, beltrama2018metalinguistic}.
At the same time, they are extremely frequent in conversational language use and are crucial for comprehending discourse.
There has been a great deal of work investigating LMs' general proficiency at function words \citep{kim2019probing} and overall sensitivity to discourse connectives \citep{pandia-etal-2021-pragmatic, beyer-etal-2021-incoherence, cong-etal-2023-investigating}. Recent work has also shown that LLMs struggle to grasp senses of discourse relations \citep{chan-etal-2024-exploring, yung-etal-2024-prompting} at a broad level. At the same time, it is unclear how well do LLMs' grasp the meaning (or senses) of discourse particles like \textit{just}---which, as we've discussed---have peculiarly interesting versatility in their semantics.

Focusing on this line of work, \textbf{this work investigates the polyfunctional discourse particle \just, which has been particularly well-studied in formal semantics} \cite[][\emph{i.a.}]{lee1987semantics, grosz2012grammar, coppock2013exclusive, beltrama2022just, thomasanddeo}.
Using data created and labeled by expert linguists, we investigate the meta-linguistic capabilities of LLMs to distinguish the nuanced senses of \just described in the formal semantics literature.
We find that while they possess basic sense distinctions, language models, especially smaller ones, struggle to fully discern the subtle differences of \just's senses, signaling the lack of a nuanced understanding of the meaning of discourse particles.\footnote{Code and data can be found here \url{https://github.com/sheffwb/IsItJUSTSemantics}}

\begin{figure*}[t]
    \centering
    \includegraphics[width=\textwidth]{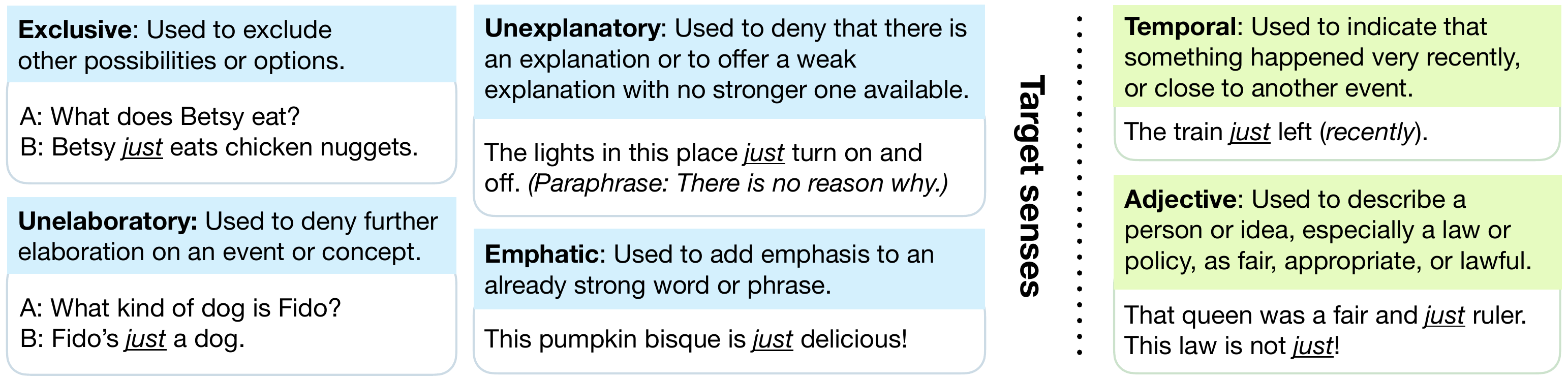}
    \vspace{-2em}
    \caption{List of \just senses in declarative sentences (with target senses in blue). Note that all senses other than the Adjective \just are discourse particle senses and function as adverbs. All examples save the Adjective ones come from \citet{warstadt2020just}.}
    \label{fig:just_senses_table}
\end{figure*}

\section{\textit{Just} and Its Semantics}
\vspace{-0.1cm}
Discourse particles, like English \just, are a class of function words that are sensitive to discourse-level contextual information. Examples include exclusive particles, such as English \just and \textit{only}, whose salient discourse function is to exclude alternatives from a contextually determined set of alternatives \cite{coppock2013exclusive}.

English \just is a good candidate to study as it (1) has been thoroughly analyzed 
and (2) has many senses.
\citet{thomasanddeo} present a unified account for all senses of the discourse particle \just,  outlining 12 senses of the word, excluding the adjective sense\footnote{The adjective sense is arguably associated with a distinct homophonous word.} (e.g. \emph{``She was a just and fair sovereign''}).
We target LLMs' ability to distinguish four of these senses that seem reasonably distinct from a semantic point of view: the exclusive \cite[][\emph{i.a.}]{coppock2013exclusive}, unelaboratory \cite{warstadt2020just}, unexplanatory \cite{wiegand2018exclusive, windhearn2021alternatives}, and emphatic \cite{lee1987semantics, beltrama2018metalinguistic, beltrama2022just}; we also use the temporal and adjective senses as controls (Figure \ref{fig:just_senses_table}).

These four senses warrant further definition.
The following examples come from \citet{warstadt2020just}.
The exclusive sense excludes other salient possibilities: In one reading of ``Betsy just eats chicken nuggets.'', the \just excludes other options of what Betsy could eat besides chicken nuggets.
The unelaboratory sense denies the need for further elaboration on an event or concept; on one reading, in response to ``What kind of dog is Fido?'' the \just in ``Fido is just a dog.'' means that Fido is simply a mutt.
The unexplanatory sense hinges on the lack of an explanation for something, and so usually has the force of adding `I don't know why'.
For example, in a haunted house someone might say ``The lights in this place just turn on and off.'', since they are not sure as to why the lights turn on and off.
The emphatic sense is used to strengthen an already extreme predicate: ``This pumpkin bisque is just delicious!'' is stronger with the \just.

The adjective sense 
is the most distinct in meaning and occurs in very different syntactic environments.
The temporal sense serves as a middle ground between the four target senses and the adjective sense: \just is still understood as a discourse particle here, but its meaning is clearly distinguishable from the other four senses.

\section{Experimental Setup}\label{sec:setup}
\vspace{-0.1cm}
\paragraph{Data}
This paper uses two sources of data to study \just: 
\textbf{(1) hand-constructed:} 90 sentences (15 of each sense) carefully created by an expert to have only one sense available for the \just of each sentence without any context; and
\textbf{(2) annotated:} 149 sentences ``in the wild'' that contain \just, taken from OpenSubtitles \cite{lison2018opensubtitles2018} and annotated by semanticists with their senses, with associated context.

The hand-constructed corpus is necessary, as clarity in the reading of the sentence is crucial for targeted metalinguistic experiments, since ambiguity can be pervasive in \justs ``in the wild''.
For example, in the sentence ``\textit{I just saw Nancy.}'', \just can either mean the seeing occurred recently (temporal reading), or only Nancy was seen (exclusive reading).\footnote{Readings are often disambiguated in speech based on the intonation of the utterance, which is not accessible to text-only models. We leave speech models for future work.} This data is created by a graduate linguist who has studied discourse particle semantics and is a native speaker of American English.

For the annotated corpus, we chose movie subtitles
over other texts as they are more conversational, and therefore more likely to contain instances of \just as a discourse particle.\footnote{Additionally, subtitles contain context, which can help disambiguate different readings for an instance of \just. However, we observe in Section \ref{sec:few-shot-results} that this has little effect on performance, further motivating our hand-constructed data.}
Our volunteer annotation team consists of two senior semanticists whose expertise is in discourse particles, and eight graduate students who have taken a graduate semantics class that extensively discussed particles. 
We collected annotations for 149 sentences, which were annotated by a variable subset of 8 annotators. When there was disagreement, we fell back on two additional senior annotators, whose labels were both considered regardless of agreement.
Table \ref{tab:subtitle_sense_dist} shows the distribution of \just senses in this subset.\footnote{There are four sentences with two occurrences of \just: in two of these, they are simply disfluencies, and so only one label is possible; in the other two, evaluating models on either occurrence did not change results.}

All sentences in both datasets have a \textbf{strong primary reading}, either by construction (in the hand-constructed corpus) or by annotator agreement (in the annotated corpus). 
While this does not rule out the possibility of multiple readings for a given sentence, strong speaker consensus on the reading of an occurrence of \just does remove more ambiguous sentences from out data. 
At the same time, this speaker consensus is a stronger signal and should be recoverable by a good model.
Examples from both datasets are in Appendix \ref{app:examples}.

\begin{table}[]
    \centering
    \small
    \begin{tabular}{c|c||c|c}
        \toprule
        \textbf{Sense} & \textbf{Count} & \textbf{Sense} & \textbf{Count} \\
        \midrule
        Exclusive & 60 & Emphatic & 21 \\
        Unelaboratory & 12 & Temporal & 33 \\
        Unexplanatory & 22 & Adjective & 1 \\
        \bottomrule
    \end{tabular}
    \vspace{-0.5em}
    \caption{Distribution of senses in annotated subtitles data. Our hand-constructed data has a balanced distribution of 15 sentences per sense.}
    \label{tab:subtitle_sense_dist}
\end{table}

\vspace{-0.1cm}
\paragraph{Models}
We use instruction-tuned models that can understand our meta-linguistic queries and evaluate diverse LLM architectures across parameter scales: \texttt{Llama-3-8b}, \texttt{Llama-3.2-1b/3b}, \texttt{Llama-3.3-70b}, \texttt{Mistral-7b-v0.3}, \texttt{OLMo-7b}, \texttt{OLMo2-7b/13b}, and  \texttt{Gemma2-2b/9b}.
All experiments were run on at most two NVIDIA A40 GPUs. Model details in Appendix \ref{app:models}.

\section{Do LLMs get nuanced \just senses?}
\label{sec:just-labeling}
\subsection{Method}
In this setting, language models are prompted to label the sense of \just in the sentence. 
The full prompt can be found in Appendix \ref{app:prompts}. 
The prompt includes both definitions and examples of each of the six senses from \citet{warstadt2020just}.
This experiment tests if the models are picking up on the information relevant to these labels even though models may not necessarily be categorizing uses of \just along the same lines as theory.

To circumvent parsing verbose generations common with prompted generation, we instead use the log probabilities of each label, conditioned on the prompt.
We take the label given the highest probability as the label assigned by the model to the sentence.
That is, the label assigned to a sentence by a model $M$ is
$\argmax_{l \in L} P_M(l | S)$
where $L$ is the set of label continuations and $S$ is the prompt including the sentence to be classified.
The conditional probability $P_M$ is calculated using \texttt{minicons}
\cite{misra2022minicons}.
Both the label continuations and the prompt are formatted to each model's chat formatting specifications.

We leverage the formatting of the sense labels directly following the in-context examples to ensure the sense labels are assigned reasonable probabilities by the model.
We directly compare the language model labels to ground-truth labels.

\subsection{Results}
\label{sec:few-shot-results}
\begin{figure}[t!]
    \centering
    \includegraphics[width=0.9\linewidth]{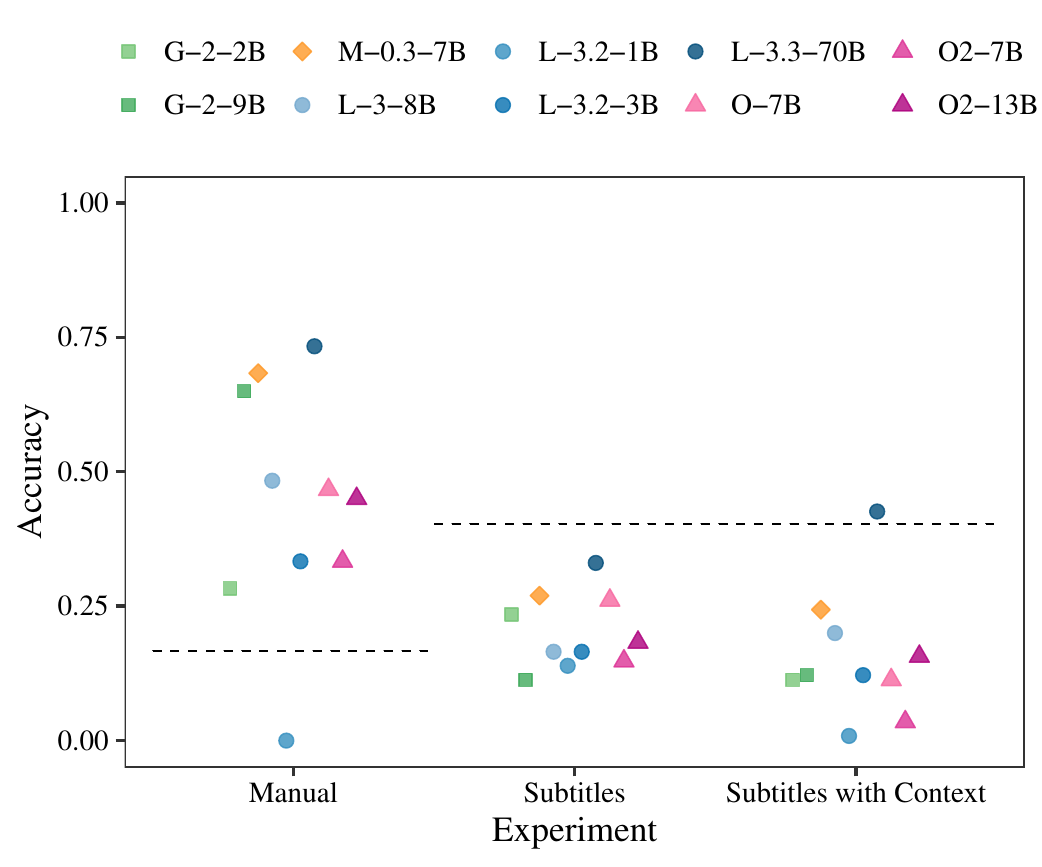}
    \vspace{-0.2cm}
    \caption{Model accuracies for the sense labeling task on our four target senses. Dashed lines show chance performance: 0.167 for hand-construction, 0.403 for subtitle data. \textbf{Model Legend:} L: Llama, G: Gemma; M: Mistral; O: OLMo.} 
    \label{fig:labeling_acc}
\end{figure}

Figure \ref{fig:labeling_acc} shows the accuracy 
for the four target senses (Exclusive, Unelaboratory, Unexplanatory, and Emphatic)\footnote{ Accuracy for all six senses is reported in Appendix \ref{app:overall-labeling-acc}.}, on three datasets: the hand-constructed data, the subtitles data alone, and the subtitles with two prior utterances as context.\footnote{We also ran this experiment with the five prior utterances and find no notable difference.}~Based on the frequency of labels, chance performance is 1/6 $\approx$ 0.167 (uniform) for the hand-constructed data and 60/149 $\approx$ 0.403 (most frequent label, Exclusive) for the subtitles data.

For the hand-constructed data, all models except \texttt{Llama-3.2-1b} substantially outperform chance.
Concerning model size, we see a substantial increase in accuracy (+0.28) from \texttt{Llama-3.2-1b} to \texttt{Gemma-2-2b}, suggesting there is a critical model size of 2B parameters required for this task (as well as for our other task in Section \ref{sec:pairwise-results}).
Additionally, we observe that the largest model, \texttt{Llama-3.3-70b}, is not performing much above the best performing mid-size models, \texttt{Mistral-7b-v0.3} and \texttt{Gemma-2-9b}, suggesting that a large model is not required for good performance.

Turning to the subtitle data, we observe a substantial drop in accuracy across all models, -0.24 on average, without context compared to the hand-constructed data.
The degradation in performance is most likely due to the subtitle sentences being more ambiguous as to what reading of \just is meant.
This indicates a notable deficit in model understanding of \just{}'s sense distinctions, since these subtitles are more naturalistic than the hand-constructed data.
Interestingly, context does \emph{not} help disambiguate the sense of \just, as we see a further decrease in model performance, -0.05 on average, when context is added, except for \texttt{Llama-3.3-70b}, which sees an increase in accuracy (+0.10), but is still at chance.
These results demonstrate an important gap in models understanding of \just senses: even when given sense definitions, they struggle to accurately predict the sense of \just in naturalistic sentences.

\section{Can LLMs distinguish \just senses?}
\label{sec:pairwise}
\vspace{-0.1cm}
\begin{figure*}[t!]
    \centering
    \includegraphics[width=0.8\textwidth]{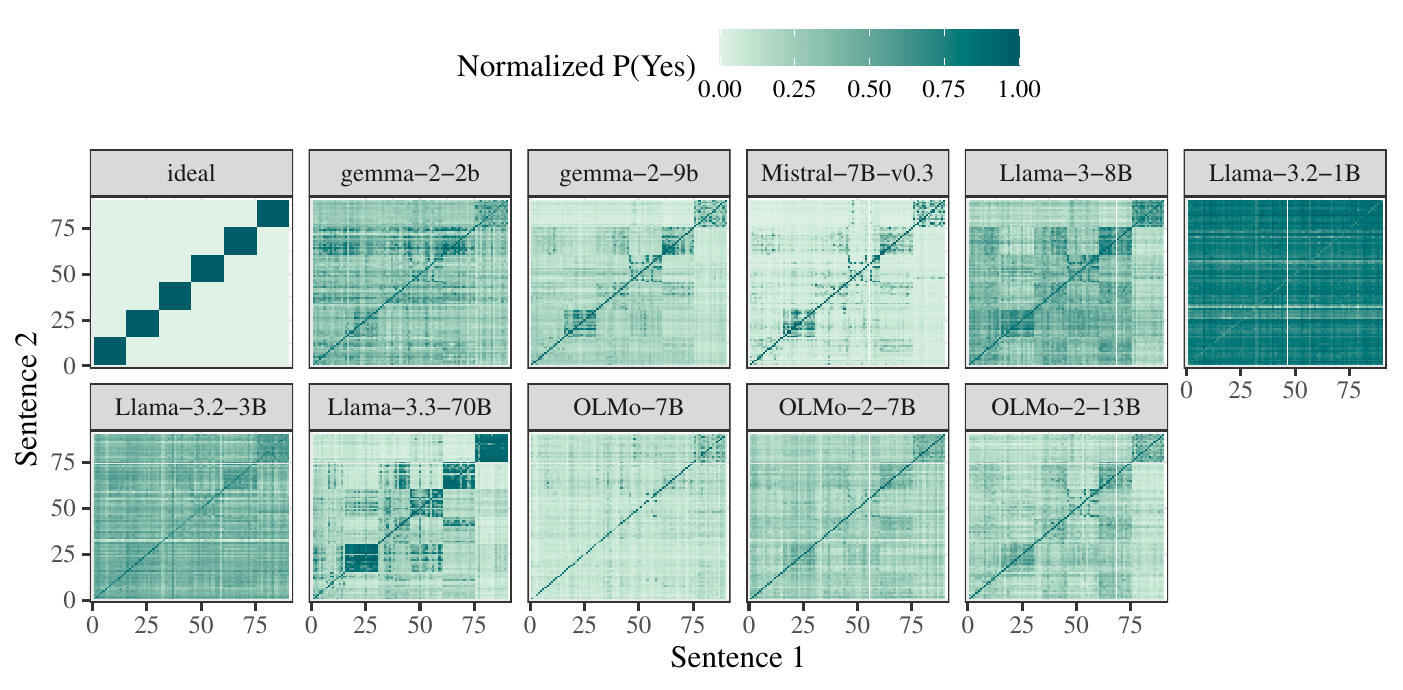}
    \vspace{-0.3cm}
    \caption{Heatmap of language model pairwise comparisons of the use of \just in the two sentences. The "ideal" heatmap shows if all sentences with the same use were judged so by the model. Senses in ascending order are: Exclusive, Unelaboratory, Unexplanatory, Emphatic, Temporal, and Adjective.}
    \label{fig:pairwise_heatmap}
\end{figure*}

\subsection{Method}
While few-shot sense labeling evaluates a model's meta-linguistic understanding of \just's senses, they come from formal linguistic theory 
and it is unclear if the differences between them are internalized in an LLM.
It is also unclear if the differences between these senses is realized in an LLM.
In order to better measure how LLMs can categorize different uses of \just, we consider model judgments on pairs of sentences, only done with the \emph{hand-constructed} data with unambiguous \just senses.

The model is given two sentences $s_i, s_j$ and prompted to answer if the \justs are used in the same way for both sentences (full prompt in Appendix \ref{app:prompts}).
Similar to the previous experiment, we compare the probabilities of the continuations ``Yes'' and ``No'' conditioned on the prompt $Z_{ij}$, which contains $s_i,s_j$. Thus, given all pairs of sentences, we define a heatmap $\mathbf{H}^M$ for each model $M$:
\vspace{-0.5em}
\[
H^M_{ij} = \log(P_M(\text{Yes} | Z_{ij})) - \log(P_M(\text{No} | Z_{ij}))
\vspace{-0.5em}
\] 
normalized to $[0,1]$ per model. 
Intuitively,
if the model judges two sentences to use \just in the same way, it will give a higher probability to ``Yes'' and a lower probability to ``No'', and vice versa.

\vspace{-0.2cm}
\paragraph{Controls} To ensure this method \emph{is} able to separate senses of words, we also perform tests for 2 words that each have multiple, clearly separate senses: 
``bat'' (2 senses) and ``bank'' (4 senses). Models show clear sense separation, verifying our method is reasonable (results in Appendix \ref{app:bat_and_bank}).

A distinct advantage of this approach is that we do not assume model knowledge of the sense labels for \just, as in the prior experiment, and instead only focus on whether they treat the meanings of just in a similar way, allowing for gradience in meaning distinctions.

\subsection{Results}
\label{sec:pairwise-results}

\paragraph{Models behave consistently and are insensitive to pair ordering.}
All models, save for \texttt{Llama-3.2-1b}, have a dark upward diagonal meaning that models see a sentence as having the same use of \just as itself (the $i=j$ diagonals); this indicates that this methodology is effective for probing model judgments on use. Additionally, the heatmaps are symmetric along the diagonals ($(i,j) \approx (j,i)$), which indicate that they are insensitive to the ordering of the sentence pairs.

\vspace{-0.1cm}
\paragraph{Models separate \just senses to some degree, but not for the nuanced target senses for particle use.}
The smallest model, \texttt{Llama-3.2-1b} is the only model to not show significant separation of the metric between pairs with the same sense and pairs with different senses ($p = .11$).
All other models show significant separation ($p < .005$).\footnote{\textit{p}-values calculated with \citet{welch1947generalization}'s t-test on the metric between pairs with the same sense but different sentences (N=1260) and pairs with different senses (N=6750).}
However, for many models the effect size is small.
Based on these distributions, all models except \texttt{Llama-3.2-1b} are able to identify sentences with the same use writ-large, although the separation is quite weak for all but the largest models (\citet{cohen-1988-statistical}'s d of 2.32 for \texttt{Llama-3.3-70b}, 1.91 for \texttt{Gemma-2-9b}, 1.66 for \texttt{Mistral-7b-v0.3}).

The strongest sense separation is for the adjective sentences, which is expected given their difference in meaning and syntactic category to the other, discourse particle senses of \just. 
Models other than \texttt{Llama-3.2-1b} also show some separation for the temporal sense.
These results show that LLMs are able to perceive different \just usages in more clearly separable senses of the word.

However, most models fail to show clear separation for the target senses, except for the Unelaboratory and Emphatic senses in \texttt{Gemma-2-9b}, \texttt{Mistral-7B-v0.3}, \texttt{Olmo-2-13b}, \texttt{Llama-3-8b}, and \texttt{Llama-3.3-70b}.
Hence, although all but the smallest models do show \emph{some} separation for \just's senses, even the largest models fail to fully capture its richness.\footnote{See Appendix \ref{app:pair-sig-tests} for details on the results in this section.} By contrast, words with relatively clear separation of senses such as \textit{bat} and \textit{bank} are more readily and consistently distinguished by models, suggesting particular gaps in the LLMs' handling of a polyfunctional particle like \textit{just}.

\vspace{-0.15cm}
\section{Conclusion}
\vspace{-0.15cm}
We find that reasonably sized language models (over 1B parameters) show some basic separation for the complex English discourse particle \just's senses, but fail to fully discern the deep subtlety of its senses with two very different prompting strategies. First, in an overt, few-shot sense-labeling setting with definitions and examples for each senses; and second in an open-ended, pairwise comparison setting allowing full freedom from \just's senses as described in formal semantics research.
This lack of nuanced sensitivity points to a gap in language model performance key for the study of discourse particles and discourse comprehension, echoing the findings of recent works \citep{chan-etal-2024-exploring,wei-etal-2024-llms, yung-etal-2024-prompting}.

\section*{Limitations}

This work looked into the English \just as a case study of LLM's metalinguistic capability to understand the semantics of discourse particles; more work needs to be performed before generalizing our findings to other discourse particles, which we leave for the future.

We have used metalinguistic prompting to analyze LLMs' understanding of \just senses. However, this class of methods has been found to underestimate LLMs' linguistic abilities, especially when compared to using direct sentence log-probabilities \citep{hu-levy-2023-prompting}. However, it is not obvious how one could analyze the nuanced distinctions in the senses of discourse particles using standard log-probability based approaches \citep[][i.a.]{warstadt-etal-2020-blimp-benchmark,hu-etal-2020-systematic, misra-etal-2023-comps}. We therefore leave this direction as an avenue for future work.

\section*{Acknowledgments}
We thank Alex Warstadt, Karen Chen, and the UT Computational Linguistics Research Seminar for their suggestions for this paper. We also thank William Thomas and all the annotators for their volunteer work in annotating the subtitle data. 
This work was partially supported by NSF grants 2104995, 2107524, and 2145479.

\bibliography{anthology,anthology_p2,custom,kanishka}

\clearpage
\appendix

\section{Dataset Examples}
\label{app:examples}
Table \ref{tab:examples} shows examples for the four primary senses from both datasets.

\section{Model Details}
\label{app:models}
Table \ref{tab:model-details} shows model details with the exact Huggingface model ID; all models are instruction-tuned and run using Huggingface's \texttt{transformers} library \cite{Wolf2019HuggingFacesTS}. All experiments are run with a temperature of 0.
\texttt{Llama-3.3-70B-Instruct} was run with 4-bit quantization.
All experiments took at most 2 hours to run for a single model, and models were run on at most 2 NVIDIA A40 GPUs.

\begin{table}[h]
    \centering
    \small
    \begin{tabular}{c|c}
         \toprule
         Huggingface Model ID & Citation \\
         \midrule
         \texttt{Llama-3.2-1B-Instruct} & \citet{llama3modelcard} \\
         \texttt{Llama-3.2-3B-Instruct} & \citet{llama3modelcard} \\
         \texttt{Meta-Llama-3-8B-Instruct} & \citet{llama3modelcard} \\
         \texttt{Llama-3.3-70B-Instruct} & \citet{llama3modelcard} \\
         \texttt{Mistral-7B-Instruct-v0.3} & \citet{jiang2023mistral} \\
         \texttt{OLMo-7B-Instruct-hf} & \citet{Groeneveld2023OLMo} \\
         \texttt{OLMo-2-1124-7B-Instruct} & \citet{olmo20242olmo2furious}\\
         \texttt{OLMo-2-1124-13B-Instruct} & \citet{olmo20242olmo2furious} \\
         \texttt{gemma-2-2b-it} & \citet{gemma_2024} \\
         \texttt{gemma-2-9b-it} & \citet{gemma_2024}\\
         \bottomrule
    \end{tabular}
    \caption{Model details. All models are instruction-tuned, and all experiments are run with a temperature of 0.}
    \label{tab:model-details}
\end{table}

\section{Overall sense labeling accuracy}
\label{app:overall-labeling-acc}
\begin{figure}[t!]
    \centering
    \includegraphics[width=0.9\linewidth]{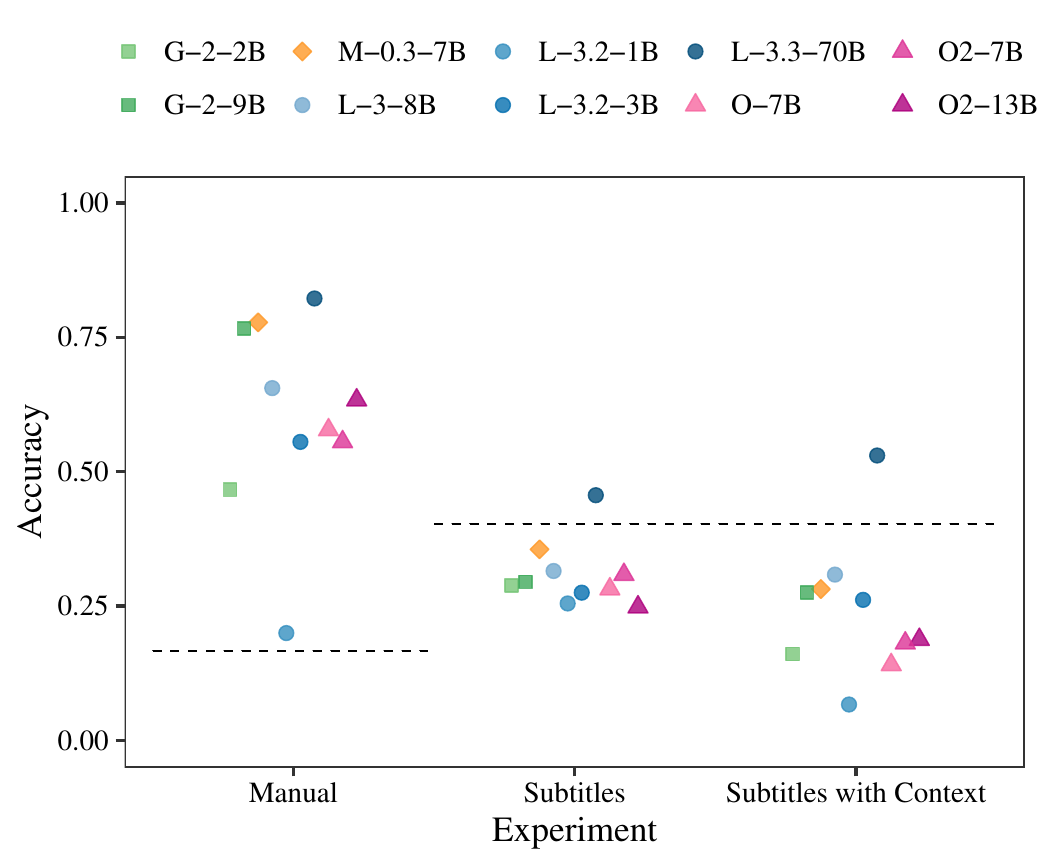}
    \vspace{-0.2cm}
    \caption{Accuracy of models for the sense labeling task on all six senses. Chance is shown with the dotted lines: 0.167 for hand-construction, 0.403 for subtitle data} 
    \label{fig:labeling_acc_app}
\end{figure}

Figure \ref{fig:labeling_acc_app} shows sense labeling accuracy on data for all six senses. Although accuracy is higher overall, there are no other notable trends, indicating that models struggle less with the temporal and adjective senses than the four target senses.

\begin{table*}[!t]
\centering
\resizebox{\textwidth}{!}{%
\begin{tabular}{@{}lll@{}}
\toprule
\textbf{Sense} & \textbf{Hand-Constructed}   & \textbf{OpenSubtitles}                  \\ \midrule
Exclusive     & The company \textbf{just} repairs existing units.      & Excuse me, judge, but this is \textbf{just} about whether or not I get bail, right? \\
Unelaboratory  & A torus is \textbf{just} a donut.    & And you're \textbf{just} in a sort of limbo. […] \\
Unexplanatory & She \textbf{just} left, out of the blue, two days ago. & Like those little doodles you \textbf{just} happened to draw?                       \\
Emphatic       & Mammoths are \textbf{just} gigantic. & […]I'm sure Ryan's gonna be \textbf{just} fine.  \\
Temporal       & I \textbf{just} received the news.   & Who were you \textbf{just} on the phone with?    \\ \bottomrule
\end{tabular}%
}
\caption{Examples for the four primary just senses from both datasets, and the temporal one for comparison.}
\label{tab:examples}
\end{table*}

\section{\textit{Bat} and \textit{Bank} pairwise sense comparisons}
\label{app:bat_and_bank}
To check that our pairwise comparison experiment is sound, we test on two additional words with clearly separable senses: ``bat'' (2 senses: a flying mammal or sports bat) and ``bank'' (4 senses: a river bank (Noun), a financial institution (Noun), to turn (Verb), or to deposit money (Verb)).

\begin{figure*}[h!]
    \centering
    \begin{subfigure}[b]{\textwidth}
        \centering
        \includegraphics[width=\textwidth]{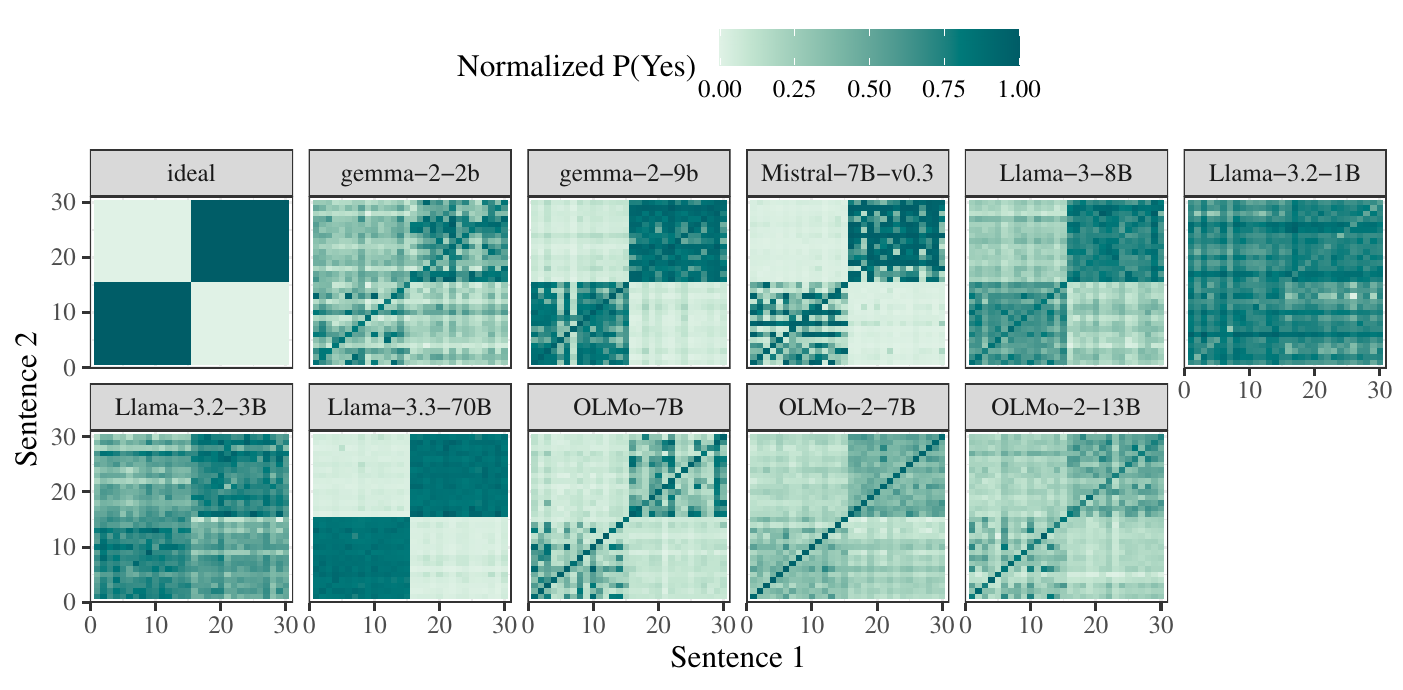}
    \end{subfigure}

    \begin{subfigure}[b]{\textwidth}
        \centering
        \includegraphics[width=\textwidth]{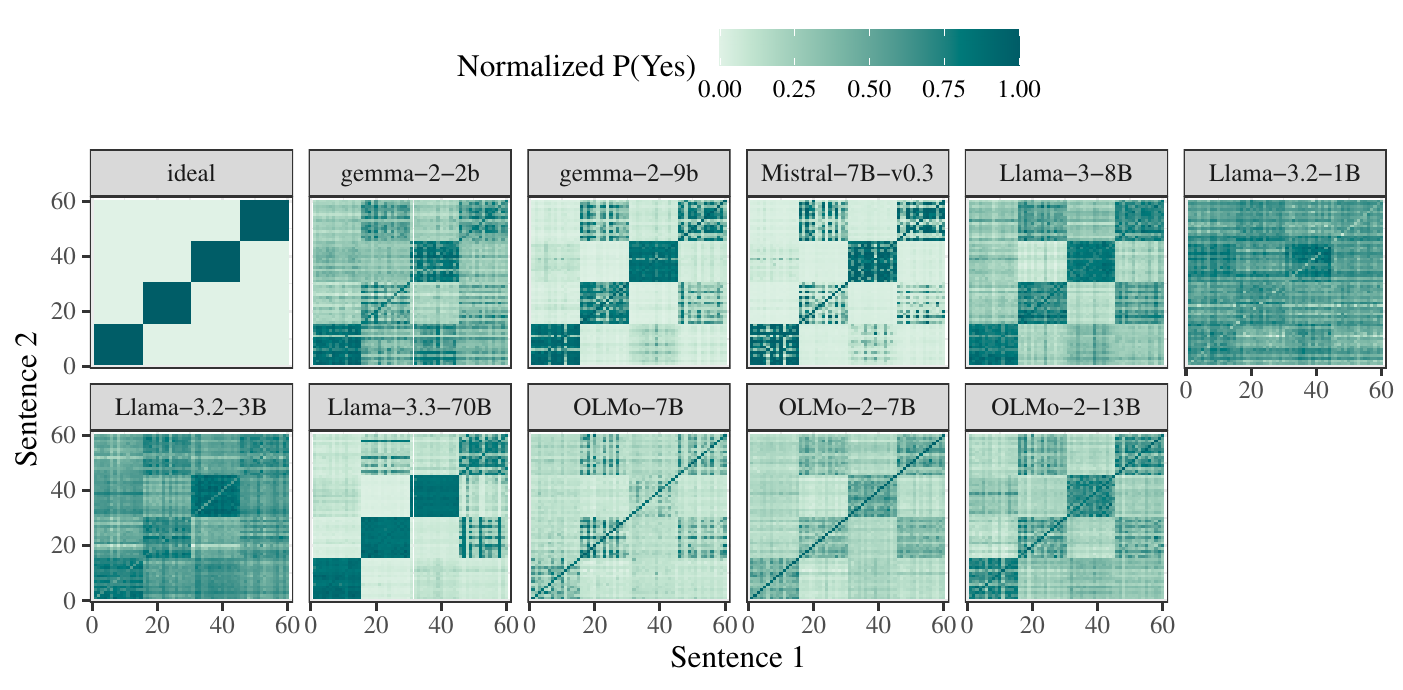}
    \end{subfigure}

    \caption{Heatmap of language model pairwise comparisons of the use of "bat" (top) and "bank" (bottom). The "ideal" heatmap shows if all sentences with the same use were judged so by the model. The senses in ascending order for (1) bat are: flying mammal, sports bat and (2) bank are: riverbank (Noun), financial institution (Noun), to turn (Verb), to deposit/keep money somewhere (Verb).}
    \label{fig:bat_bank_heatmaps}

\end{figure*}

Heatmaps for model pairwise comparisons of \textit{bat} and \textit{bank} are shown in Figure \ref{fig:bat_bank_heatmaps}. Each includes an idealized heatmap, where only pairs with the same sense are given a score of 1 and the rest 0. 

For both ``bat'' and ``bank'' we see clear separation of senses, as seen by the squares along the diagonal for all models, with \texttt{Llama-3.3-70b} amd \texttt{Gemma-2-9b} being closest to the ideal matrix.
This indicates that this method is viable for testing language model ability to separate senses using this pairwise comparison methodology.

Interestingly, we do see notable confusion between the financial institution sense (Noun) and the deposit money sense (Verb) with the hot spot off the diagonal, showing that the models are focusing on meaning and not syntactic differences, as focus on syntax would be shown by hot spots between the two verbs and/or nouns.

\begin{table}[t]
    \centering
    \small
    \begin{tabular}{c|c|l}
        \toprule
        \textbf{Distribution $\mu\pm\sigma$} & \textbf{\textit{p}-value} & \textbf{Model} \\
        \midrule
        -0.016±0.118 &  <0.0001 & \texttt{Llama-3.2-1b} \\
        -0.008±0.087 & <0.0001 & \texttt{Llama-3.2-3b} \\
        0.011±0.132 & <0.0001 & \texttt{Llama-3.3-70b} \\
        0.027±0.130 & <0.0001 & \texttt{Meta-Llama-3-8b} \\
        0.008±0.100 & <0.0001 & \texttt{Mistral-7b-v0.3} \\
        0.003±0.100 & 0.003 & \texttt{OLMo-2-1124-13b} \\
        -0.008±0.086 & <0.0001 & \texttt{OLMo-2-1124-7b} \\
        -0.010±0.067 & <0.0001 & \texttt{OLMo-7b} \\
        -0.039±0.163 & <0.0001 & \texttt{gemma-2-2b} \\
        0.027±0.073 & <0.0001 & \texttt{gemma-2-9b} \\
        \bottomrule
    \end{tabular}
    \caption{\textit{p}-values for a two-sided one-sample t-test on, and distributions for, $(i,j) - (j,i), i \leq j$}
    \label{tab:flip_values}
\end{table}

\section{Significance Tests for Pairwise Experiments}
\label{app:pair-sig-tests}

Table \ref{tab:flip_values} shows the distribution of the sentence pair metrics $(i,j)-(j,i)$ for $i\leq{}j$, as well as the \textit{p}-value for a two-sided, one sample t-test for the mean being different from 0 (8100 total sentence pairs).
This tests if models exhibit ordering preferences for senses of \just.
Although all means are significantly different from 0, they are never farther than .039 (\texttt{Gemma2-2b}), and all but three are less than 0.016 from 0, indicating the effect of sentence ordering is minimal for the pairwise experiment.
Hence, we conclude the models have minimal ordering bias in the pairwise experiment.

Additionally, visually we can observe in Figure \ref{fig:pairwise_heatmap} no clear ordering preference for any particular sense, which would be indicated by a strong, dark horizontal or vertical band in the heatmap.
We see a \emph{possible} such band for \texttt{Gemma2-2b} with the horizontal temporal pairings, indicating the this model is slightly more likely to force non-temporal readings for a sentence pair ($s_i, s_j$) if $s_i$'s \just isn't temporal.
These absence of such bands everywhere else demonstrates the lack of ordering preferences in these models for this task.

Table \ref{tab:sense-separation} contains the distributions of the metric for the pairs with the same sense (but not the exact same sentence $i=j$) and pairs with different senses. 
\textit{p}-values calculated with a one-sided \citet{welch1947generalization}'s t-test on the metric between pairs with the same sense but different sentences (1260 instances) greater than pairs with different senses (6750 instances).
The effect size is calculated using \citet{cohen-1988-statistical}'s d.
Therefore greater effect size means a better separation of senses.

\begin{table*}[]
    \centering
    \small
    \begin{tabular}{lccc|l}
        \toprule
        \textbf{Same Sense $\mu\pm\sigma$} & \textbf{Different Sense $\mu\pm\sigma$} & \textbf{\textit{p}-value} & \textbf{Effect Size} & \textbf{Model} \\
        \midrule
        0.774$\pm$0.114  & 0.770$\pm$0.102  & 0.114  & 0.040    & \texttt{Llama3.2 1B} \\
        0.477$\pm$0.122  & 0.398$\pm$0.104  & <0.0001  & 0.729  & \texttt{Llama3.2 3B} \\
        0.594$\pm$0.287  & 0.191$\pm$0.143 & <0.0001 & 2.319   & \texttt{Llama3.3 70B} \\
        0.548$\pm$0.142  & 0.322$\pm$0.139  & <0.0001 & 1.615   & \texttt{Llama3 8B} \\
        0.265$\pm$0.236  & 0.076$\pm$0.071  & <0.0001 & 1.664    & \texttt{Mistral-7B-v0.3} \\
        0.356$\pm$0.152  & 0.195$\pm$0.096  & <0.0001 & 1.516    & \texttt{OLMo2 13B} \\
        0.323$\pm$0.121  & 0.226$\pm$0.080  & <0.0001 & 1.105   & \texttt{OLMo2 7B} \\
        0.162$\pm$0.126  & 0.117$\pm$0.059  & <0.0001  & 0.602   & \texttt{OLMo 7B} \\
        0.437$\pm$0.172  & 0.326$\pm$0.134  & <0.0001  & 0.785   & \texttt{Gemma2 2b} \\
        0.360$\pm$0.191  & 0.156$\pm$0.082  & <0.0001 & 1.905   & \texttt{Gemma2 9b} \\
        \bottomrule
    \end{tabular}
    \caption{Comparison of model separation for pairs with the same sense versus different senses, excluding pairs with the same sentence repeated. \textit{p}-values are calculated using  \citet{welch1947generalization}'s t-test, with same sense being greater than different sense. Effect size is calculated using \citet{cohen-1988-statistical}'s d.}
    \label{tab:sense-separation}
\end{table*}

\section{Prompts}
\label{app:prompts}
\begin{figure*}
    \centering
    \includegraphics[width=\textwidth]{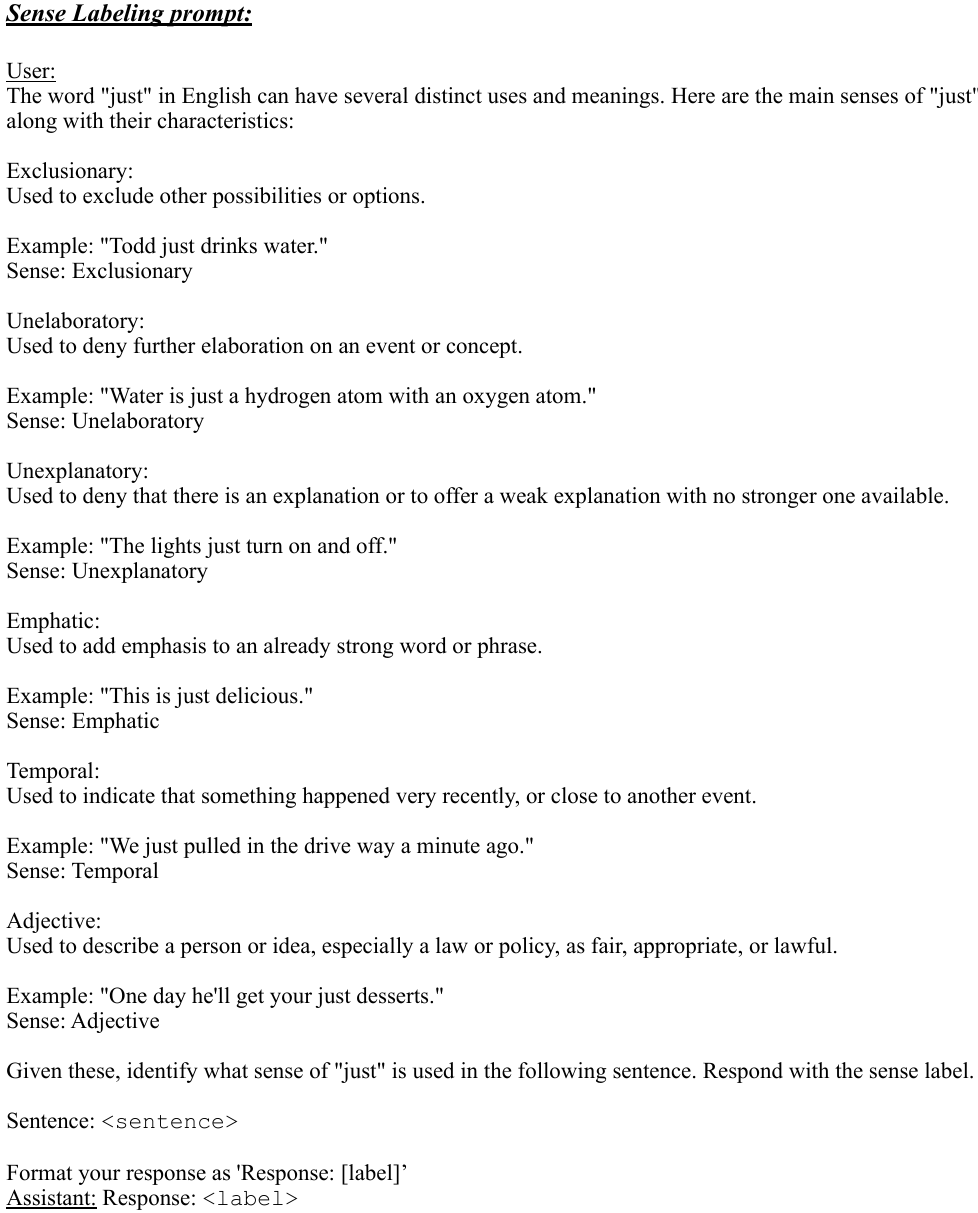}
    \caption{Prompt for the "just" sense labeling task. When context is included for the subtitle data, the prompt is slightly altered to: "Identify what sense of "just" is used in the last sentence of the following passage".}
    \label{fig:sense_label_prompt}
\end{figure*}

\begin{figure*}
    \centering
    \includegraphics[width=\textwidth]{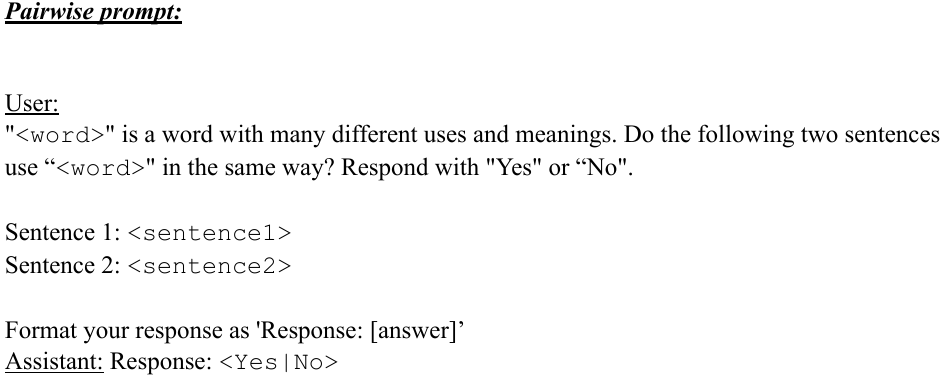}
    \caption{Prompt for the pairwise sense comparisons.}
    \label{fig:pairwise_prompt}
\end{figure*}

The prompts used for the experiments in Sections \ref{sec:just-labeling} and \ref{sec:pairwise} can be found in Figures \ref{fig:sense_label_prompt} and \ref{fig:pairwise_prompt}, respectively.
For Section \ref{sec:just-labeling}, the prompt includes shorter sense definitions to match what human annotators were given.

\section{Annotation Interface}
\label{app:ann-interface}
\begin{figure*}
    \centering
    \includegraphics[width=\textwidth]{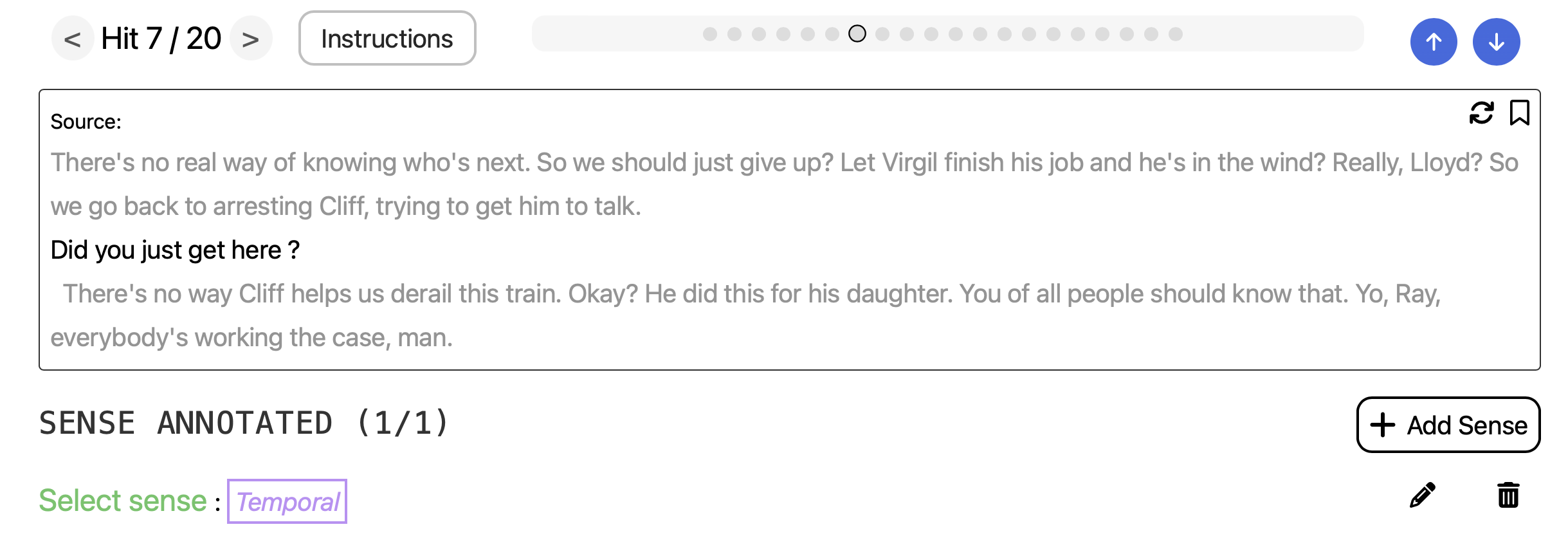}
    \caption{Annotation interface for labeling subtitle data.}
    \label{fig:annotation-interface}
\end{figure*}

Figure \ref{fig:annotation-interface} shows an example of the interface annotators used to label the subtitle data. Users were given 5 sentences of context, and selected one of the sense labels. They could also include comments.
Annotators were trained in an in-person session with one of the authors.

\end{document}